\documentclass[sigconf]{acmart}
\AtBeginDocument{%
  }

\setcopyright{acmlicensed}
\acmDOI{XXXXXXX.XXXXXXX}

\usepackage{arydshln}
\usepackage{amsmath,mathtools,bbm}
\usepackage{subcaption}
\usepackage{multirow}

\usepackage{algorithm}
\usepackage{algpseudocode}

\usepackage{graphicx}
\usepackage{xcolor}

\begin{document}

\definecolor{ForestGreen}{RGB}{34,139,34}

\newcommand{\think}[1]{\textcolor{blue}{\texttt{<think>}} #1 \textcolor{blue}{\texttt{</think>}}}
\newcommand{\search}[1]{\textcolor{cyan}{\texttt{<search>}} #1 \textcolor{cyan}{\texttt{</search>}}}
\newcommand{\info}[1]{\textcolor{ForestGreen}{\texttt{<information>}} #1 \textcolor{ForestGreen}{\texttt{</information>}}}
\newcommand{\more}[1]{\textcolor{orange}{\texttt{<more info>}} #1 \textcolor{orange}{\texttt{</more info>}}}
\newcommand{\answer}[1]{\textcolor{purple}{\texttt{<answer>}} #1 \textcolor{purple}{\texttt{</answer>}}}

\newcommand{\helia}[1]{\textcolor{red}{helia: #1}}
\title{Cost-Aware Retrieval-Augmentation Reasoning Models with Adaptive Retrieval Depth}


\author{Helia Hashemi}
\affiliation{%
  \institution{Microsoft}
  \city{Cambridge}
  \state{MA}
  \country{United States}}
\email{heliahashemi@microsoft.com}

\author{Victor R\"{u}hle}
\affiliation{%
  \institution{Microsoft}
  \city{Cambridge}
  \country{United Kingdom}}
\email{virueh@microsoft.com}

\author{Saravan Rajmohan}
\affiliation{%
  \institution{Microsoft}
  \city{Redmond}
  \state{WA}
  \country{United States}}
\email{rajmohan@microsoft.com}








\begin{abstract}
Reasoning models have gained significant attention due to their strong performance, particularly when enhanced with retrieval augmentation. However, these models often incur high computational costs, as both retrieval and reasoning tokens contribute substantially to the overall resource usage. In this work, we make the following contributions: (1) we propose a retrieval-augmented reasoning model that dynamically adjusts the length of the retrieved document list based on the query and retrieval results; (2) we develop a cost-aware advantage function for training of efficient retrieval-augmented reasoning models through reinforcement learning; and (3) we explore both memory- and latency-bound implementations of the proposed cost-aware framework for both proximal and group relative policy optimization algorithms. We evaluate our approach on seven public question answering datasets and demonstrate significant efficiency gains, without compromising effectiveness. In fact, we observed that the model latency decreases by $\sim$16-20\% across datasets, while its effectiveness increases by $\sim$5\% on average, in terms of exact match. 
\end{abstract}


\maketitle

\section{Introduction}

Retrieval-augmented generation (RAG) has been respected as a standard technique for improving accuracy and factuality of large language models (LLMs) in addition to providing access to up-to-date information that was not available in their training data \cite{RAG,REML}. Recently, LLMs capable of reasoning have proven successful in interacting with search engines to acquire knowledge during their reasoning process. Search-o1 \cite{searcho1} and Search-R1 \cite{jin2025searchr1} are examples of such models. Despite their success, the types of interactions they have with search engines are limited and insufficient. In more detail, these models can only submit a query to a search engine and get a constant number of documents in response. However, some queries may require synthesis of more documents, or the LLM may need to ask for deeper ranked documents since the presented top retrieved documents are not relevant. One may argue that we can always feed a large enough number of documents to the LLM in response to each query to mitigate this issue. Although increasing context size and processing more documents is likely to improve the model's average performance, it would come at a high computational cost, high latency, and may even result to some other issues, such as suffering from the loss-in-the-middle phenomenon \cite{loss-in-the-middle}.

To address these issues, we extend the recent open-source Search-R1 model \cite{jin2025searchr1} and develop reinforcement learning-based optimization methods that enable LLMs to learn more dynamic interactions with search engines, called Dynamic Search-R1. We choose Search-R1 as a representative retrieval-augmented reasoning model due to its popularity in the literature \cite{liu:2025, zhang:2025,xue:2025,chen:2025}, but our work is generic enough to be applied to any retrieval-augmented reasoning model. In Dynamic Search-R1, the model not only generates reasoning tokens within \think{and}, but also generates queries that can be submitted to the search engine in \search{query} that returns the top retrieved document and asks for $k$ more retrieved documents using \more{$k$}, where $k$ can be any positive integer number. The model is trained based on an outcome-based reward function that solely examines the final answer and basically learns interactions with the search engines through trial and error within a predefined rollout budget. In order to prevent the model from repeatedly asking for more information when is not necessary and to improve the efficiency of the model, we propose a cost-aware variant of Dynamic Search-R1 for both Group Relative Policy Optimization (GRPO) and Proximal Policy Optimization (PPO) algorithms. 

We provide two implementations of the cost penalization functions in our study. We study a memory-bound penalizer that focuses on minimizing the number of tokens produced by the model, including both the generated tokens and the ones returned by the search engine in response to the generated queries. Even though reducing the number of tokens generally reduces latency in LLMs, this assumption does not necessarily hold for Dynamic Search-R1. The reason is that the tokens in its reasoning chain can come from both LLM generation and the retrieved documents and there is a substantial gap between them from an efficiency perspective: we observe that the latency for token generation is over 600\% more than the latency for token encoding in our experiments. This observation has motivated us to further study a latency-bound cost function that distinguishes between generated and retrieved tokens during reasoning.

Experiments on seven diverse general and multi-hop question answering datasets suggest that Dynamic Search-R1 not only outperforms Search-R1 in terms of answer quality, but also performs more efficiently when trained with a cost-aware advantage function during the reinforcement learning (RL) process. We demonstrate that this observation holds for LLMs of different sizes (i.e., 3B and 7B parameters). We further provide insights into the impact of the hyper-parameter value that controls the cost term in RL training.

\section{Related Work}
This section reviews prior work on retrieval-augmented generation and length penalization in text generation--the two areas mostly related to our contribution on cost-aware retrieval-augmented reasoning.

\subsection{Retrieval-Augmented Generation}
Integrating search engines with LLMs has become a common strategy for providing external information and partially addressing the well-known hallucination issue. There are two primary ways to integrate search
 engines with LLMs: (1) retrieval-augmented generation (RAG) \cite{gao2023retrieval} and (2)
 treating the search engines as tools \cite{schick2023toolformer}. RAG 
 \cite{lewis2020retrieval, yue2024inference, xiong2025rag} typically involves a pipeline where a query triggers a retrieval step, and the retrieved content is appended to the query before being processed by the LLM. This common approach is called in-context augmentation. Fusion-in-decoder models \cite{izacard-grave-2021-leveraging,FiD-Light} provide another implementation of retrieval augmentation by encoding passages independently with encoder-decoder LLMs. The ``search-as-a-tool'' paradigm enables LLMs to engage directly with search engines through prompting or fine-tuning. Methods such as IRCoT \cite{trivedi2022interleaving} and ReAct \cite{yao2023react} use prompting to support step-by-step reasoning and iterative search interactions, whereas Toolformer \cite{schick2023toolformer} enhances this capability via supervised fine-tuning. However, these techniques are heavily reliant on large-scale, high-quality labeled data, which is challenging to acquire. Recent studies \cite{guo2025deepseek} suggest that reinforcement learning can help LLMs develop advanced reasoning abilities using only outcome-based feedback. Application of this work in the context of search engine interaction has been studied by \citet{jin2025searchr1}. They have introduced Search-R1, which relies on RL for incorporating search engines into a reasoning model. One key distinction of this work compares to others is that the model able to reason and query the search engine in an iterative manner for a reasoning task. The model and optimization that we proposed in this work is built on Search-R1, which is the most competitive baseline for this work. None of the reviewed literature performs dynamic augmentation as proposed in this work.\footnote{Note that there is a recent paper called `Dynamic RAG' \cite{su-etal-2024-dragin} which focuses on exploring `when to retrieve' in RAG, which is fundamentally different from this work and is just a name similarity.}

\subsection{Text Generation with Length Penalization}
Controlling the output length of large language models (LLMs) is crucial for many generation tasks. Existing methods approach this through architectural tweaks—like adjusting positional encodings for fixed-length outputs \cite{Butcher_2025}, modifying training objectives to enforce length constraints \cite{jie2023,singhal2024l}, or using instruction-style data labeled with target lengths \cite{yuan2024}. These efforts typically fall into two categories: reducing verbosity or achieving strict length limits and token-level precision. However, most prior work centers on general-purpose generation or instruction-following, often overlooking efficiency concerns. L1 \cite{aggarwal2025l1controllinglongreasoning} and S1 \cite{muennighoff2025s1simpletesttimescaling} models are among few papers that focus on controlling length in reasoning-oriented models. S1 attempts to regulate output length by requiring the model to omit specific tokens (e.g., ``Wait'' or ``Final Answer'') when the output is either too short or too long. However, this rigid and manually designed method significantly harms performance relative to the base model. L1 is a reasoning language model that produces output that satisfies a length constraint given in its prompt. L1 has been trained using Length Controlled Policy Optimization (LCPO),
 a simple reinforcement learning method that optimizes for accuracy and adherence to user-specified length constraints. It is important to note that both of these methods give the users a tool to control output length at the cost of performance rather than searching for optimal number of tokens without performance loss. Also, none of these methods are studied in the context of retrieval augmentation. Although, they could be easily adopted, it is not clear how many tokens should be fixed for the limit as the retrieval depth could be different. There are efficient retrieval-augmented generation models, but they mostly focus on context compression \cite{verma2024contextualcompressionretrievalaugmentedgeneration,FiD-Light} or retrieval acceleration through specific hardware design \cite{xRAG}, which are all orthogonal to our work.

\section{Cost-Aware Dynamic Retrieval-Augmented Reasoning}
Enabling reasoning models to acquire knowledge through retrieval has proven to be promising for knowledge-intensive tasks, especially where domain-specific or up-to-date information is required that is not accurately captured by the LLM during its training. Although retrieval-augmented reasoning models could be powerful, they can be quite costly as well. The reason is that they may generate many reasoning tokens and consume many tokens from retrieved documents that can add up and lead to inefficient and expensive models. We argue that we should give models the power of searching for queries, and even beyond that, by performing more actions on the search results, such as asking for more information (e.g., similar to going to the next page in the search engine result page). On the other hand, the reasoning models should be aware of the cost associated with each of these actions. Based on this motivation, we in the following introduce a cost-aware approach for training retrieval-augmented reasoning models through reinforcement learning.

\subsection{Background: Training Retrieval-Augmented Reasoning Models with Reinforcement Learning}
Reasoning models that are capable of acquiring knowledge through retrieval can be formulated using reinforcement learning as follows:
\begin{align}\label{eq:rl-retriever}
    \max_{\pi_\theta}& ~\mathbb{E}_{x \sim \mathcal{D}, y \sim \pi_{\theta}(\cdot \mid x; \mathcal{R})} 
\left[ r_{\phi}(x, y) \right] \nonumber \\
&- \beta \mathbb{D}_{\text{KL}} \left[ \pi_{\theta}(y \mid x; \mathcal{R}) \,||\, \pi_{\text{ref}}(y \mid x; \mathcal{R}) \right],
\end{align}
where $\pi_{\theta}$ is the policy LLM, $\pi_{\text{ref}}$ is the reference LLM, $r_{\phi}$ is the reward function, $\mathcal{R}$ is the retrieval model, and $\mathbb{D}_{\text{KL}}$ denotes the KL-divergence between the given two distributions.
\( x \) denotes input samples drawn from the dataset \( \mathcal{D} \), and \( y \) represent the generated outputs interleaved with search engine calling results, sampled from the reference policy \( \pi_{\text{ref}}(y \mid x) \).
This formulation explicitly incorporates retrieval interleaved reasoning via $\pi_{\theta}(\cdot \mid x; \mathcal{R})$.
This enables more effective reasoning where access to external knowledge is required. 

This generic RL approach can be implemented using different policy gradient methods. We build our experiments on the recent Search-R1 \cite{jin2025searchr1} method that implements this RL framework using Proximal Policy Optimization (PPO) \citep{schulman2017proximal} and Group Relative Policy Optimization (GRPO) \citep{shao2024deepseekmath,guo2025deepseek}. The objective functions used in Search-R1 are presented in Table~\ref{tab:loss}. Additionally, Search-R1 masks the tokens from the retrieval results to only update the model parameters for the tokens it generated. These formulations are used as the foundation for our implementation.

\begin{table*}[h]
    \centering
        \caption{PPO and GRPO objectives for retrieval-augmented reasoning models, as defined by Search-R1 \cite{jin2025searchr1}.}\label{tab:loss}
    \resizebox{\textwidth}{!}{
    \begin{tabular}{p{\textwidth}}
        \hline
        \multicolumn{1}{c}{\textbf{PPO}}\\\hdashline
        {\footnotesize
        \begin{equation*}
            \mathbb{E}_{x \sim \mathcal{D}, y \sim \pi_{\text{old}}( \cdot| x; \mathcal{R})} \Bigg[ \frac{1}{\sum_{t=1}^{|y|} I(y_t)} \sum_{t=1: I(y_t)=1}^{|y|} \min \left( \frac{\pi_{\theta}(y_t | x, y_{<t}; \mathcal{R})}{\pi_{\text{old}}(y_t | x, y_{<t}; \mathcal{R})} A_t, \text{clip} \left( \frac{\pi_{\theta}(y_t | x, y_{<t}; \mathcal{R})}{\pi_{\text{old}}(y_t | x, y_{<t}; \mathcal{R})}, 1 - \epsilon, 1 + \epsilon \right) A_t  \right) \Bigg]
        \end{equation*}}
        \\
        \hline
        \multicolumn{1}{c}{\textbf{GRPO}}\\\hdashline
        {\footnotesize
        \begin{align*}\label{eq:grpo}
            \mathbb{E}_{x \sim \mathcal{D}, \{ y_i \}_{i=1}^{G} \sim \pi_{\text{old}}( \cdot| x; \mathcal{R})}&
            \Bigg[\frac{1}{G} \sum_{i=1}^{G} \frac{1}{\sum_{t=1}^{|y_i|}  I(y_{i,t})} \sum_{t=1: I(y_{i,t})=1}^{|y_i|} 
            \min \Bigg( \frac{\pi_{\theta}(y_{i,t} | x, y_{i,<t}; \mathcal{R})}{\pi_{\text{old}}(y_{i,t} | x, y_{i,<t}; \mathcal{R})} {A}_{i,t},  \nonumber \\[8pt] 
             &\hspace{120pt} \text{clip} \Bigg( \frac{\pi_{\theta}(y_{i,t} | x, y_{i,<t}; \mathcal{R})}{\pi_{\text{old}}(y_{i,t} | x, y_{i,<t}; \mathcal{R})}, 1 - \epsilon, 1 + \epsilon \Bigg) {A}_{i,t} \Bigg) - \beta \mathbb{D}_{KL} \left[ \pi_{\theta} || \pi_{\text{ref}} \right] \Bigg] 
        \end{align*}
        }\\
        \hline
    \end{tabular}
    }
\end{table*}

\subsection{Retrieval-Augmented Reasoning with Adaptive Retrieval Depth}
Search-R1 and other retrieval-augmented reasoning models, such as OpenAI's Deep Research\footnote{\url{https://openai.com/index/introducing-deep-research/}} or Search-o1 \cite{searcho1}, assume that the LLM always has access to a constant number of retrieval results per query. Therefore, if the provided search results are not sufficient, the mode cannot simply ask for more lower ranked documents in the search results, and its only choice is to reformulate its query. We enable the model to ask for more documents when needed by using the following prompt:

{\footnotesize
\begin{quote}
    Answer the given question. \
        You must start the output by generating reasoning tokens inside \think{and} and every time you get new information. \
        After reasoning, if you lack some knowledge, you can call a search engine by \search{query}, and it will return the top searched result between \info{and}. \
        After each search you can optionally ask for more documents by \more{num\_docs} where num\_docs is a positive integer indicating the number of additional requested documents. Additional retrieved documents will be provided between \info{and}.
        You can search as many times as you want. \
        If you find that no further external knowledge is needed, you can directly provide the answer inside \answer{and} without detailed illustrations. For example, \answer{xxx}. Question: \textcolor{red}{\texttt{\{question\}}}.\\
\end{quote}
}
\noindent where \textcolor{red}{\texttt{question}} is replaced with the specific question during training and inference.

In addition, we use Algorithm~\ref{alg:llm_search} for producing the rollouts for both training and inference. Note that all retrieved results for both new queries and additional requested documents are masked using the identity function $I$ in the PPO and GRPO objectives. See the equations in Table~\ref{tab:loss}.

\begin{algorithm*}[t]
\caption{LLM Response Rollout with Multi-Turn Search Engine Calls and Adaptive Retrieval Depth}
\label{alg:llm_search}
\begin{algorithmic}[1]
\Require Input query \( x \), policy model \( \pi_{\theta} \), search engine \( \mathcal{R} \), maximum action budget \( B \).
\Ensure Final response \( y \).

\State Initialize rollout sequence \( y \gets \emptyset \)
\State Initialize action count \( b \gets 0 \)

\While{\( b < B \)}
    \State Initialize current action LLM rollout sequence \( y_b \gets \emptyset \) 
    \While{True}
    \State Generate response token \( y_t \sim \pi_{\theta}(\cdot \mid x, y + y_b) \)
    \State Append \( y_t \) to rollout sequence \( y_b \gets y_b + y_t \)
    \If{\( y_t \) in [\textcolor{cyan}{\texttt{</search>}}, \textcolor{orange}{\texttt{</more info>}}, \textcolor{purple}{\texttt{</answer>}},  \texttt{<eos>}]}
        break
    \EndIf
    \EndWhile

    \State \( y  \gets  y + y_b \)
    \If{\search{} detected in \( y_b \)}
        \State Extract search query \( q \gets \text{Parse}(y_b, \search{} ) \)
        \State Retrieve search results \( d = \mathcal{R}(q) \)
        \State Insert $d$ into rollout \( y  \gets  y + \info{d}  \)
    \ElsIf{\more{} detected in \( y_b \)}
        \State Extract the number of requested documents \( k \gets \text{Parse}(y_b, \more{~} ) \)
        \State Fetch the next $k$ documents \( d \) from the cached retrieval results
        \State Insert $d$ into rollout \( y  \gets  y + \info{d}  \)
    \ElsIf{\answer{} detected in \( y_b \)}
        \State \textbf{return} final generated response \( y \)
    \Else
        \State Ask for rethink \( y  \gets  y + \) ``My action is not correct. Let me rethink.''
    \EndIf

    \State Increment action count \( b \gets b + 1 \)
\EndWhile

\State \textbf{return} final generated response \( y \)
\end{algorithmic}
\end{algorithm*}

\subsection{Cost-Aware Advantage Function for Improved Efficiency}
In this section, we describe our approach for cost-aware reasoning models that are capable of interactive multi-turn retrieval. We modify the advantage function used in RL optimization methods. 

\textbf{Cost-Aware Advantage Function for GRPO.} In GRPO, the advantage function is defined as a group relative normalized function applied to the reward value obtained for each output. Let $o_1, o_2, \cdots, o_G$ denote a set of $G$ outputs sampled from the LLM being trained. We define the cost-aware advantage function for output $o_i$ as follows:
\begin{align}
    A_i = &\frac{r_{i} - \text{mean}(\{r_1, r_2, \cdots, r_G\})}{\text{std}(\{r_1, r_2, \cdots, r_G\})} \nonumber\\
    &- \alpha \left(\frac{c_{i} - \text{mean}(\{c_1, c_2, \cdots, c_G\})}{\text{std}(\{c_1, c_2, \cdots, c_G\})}\right)
\end{align}
where $r_i$ and $c_i$ denote the reward (benefit) and the cost associated to output $o_i$. The hyperparameter $\alpha$ controls the weight of cost during optimization. A nice property of this group relative normalization for cost is that training queries are of different difficulty levels, one may not need substantial reasoning, while others may need substantial step-by-step retrieval and reasoning. Therefore, absolute cost-aware methods would over-penalize the model for complex queries. However, the proposed cost-aware advantage function normalizes the cost of $G$ sample outputs for the same query, ensuring that only an output with a high cost compared to its alternative solutions for the same query would be penalized. 

Following \cite{jin2025searchr1}, we use exact match of the final answer as a simple outcome-based reward value. The cost $c_i$ can be computed in different ways. We adopt the following two cost functions:
\begin{itemize}
    \item \textbf{Memory-bound Cost:} Given one input, the GPU memory required for generating two outputs would depend on the number of additional tokens generated and fed to the model. Therefore, the length of model outputs (i.e., the number of tokens, including the tokens from retrieved documents) is considered as a memory-bound cost function.

    \item \textbf{Latency-bound Cost:} This cost function focuses on model's latency for producing the output, using the following function:
    \begin{equation}
        c_i = \sum_{t=1}^{|y_i|}{I(y_{i, t})} \cdot c_g + \sum_{t=1}^{|y_i|}{(1-I(y_{i, t}))} \cdot c_e
        \label{eq:latency_cost}
    \end{equation}
    where $I$ is an identity function indicating whether a token is generated by the LLM or is provided by the search engine. $c_g$ and $c_e$ denote the average latency for each generated token and each provided token for encoding (i.e., each token from retrieval results). The reason for this distinction is that latency for generated token is much higher than each encoded token. We ran multiple experiments with different input and output sizes to profile latency introduced by input token consumption and output token generation. In the experiments, we used one A100 GPU with 80GB memory and Qwen-2.5-7B \cite{qwen2025qwen25technicalreport} as a representative model. Experiments suggest that representing each input token takes 0.0568 milliseconds on average, while generating each output token on average requires 0.4098 milliseconds. Therefore, generating output tokens is 621\% slower than consuming input tokens. Of course these numbers depend on the GPU and the consumed model and the cost parameters mentioned above should be assigned according to the specific experiment design. That said, we observe relatively similar behavior across LLMs and GPUs. 
\end{itemize}

\textbf{Cost-Aware Advantage Function for PPO.} The advantage function for PPO is defined using a generalized advantage estimation (GAE) \cite{GAE} as follows:
\begin{equation}
    A_t = -V_\phi(x, y_{<t}; \mathcal{R}) + G_t
\end{equation}
where $v_\phi$ is a value model that estimates the expected value function of incomplete responses under an active policy and $G_t=\sum_{t'=t}^{|y|}{\gamma^{t'-t}r_{t'}}$ is the empirical return estimated by GAE. The value model $v_\phi$ shares the same architecture as the policy model, while the regression head can be applied on any response token. To construct token-level rewards $r_t$ that guide the RL training, the outcome-based reward $r$ discounted by the cost is applied only to the last token, and each token is penalized by a Kullback-Leibler term. Formally, the token-level reward $r_t$ for each response token $y_t$ is defined as follows:
\begin{align}
    r_t = -\beta \log& \frac{\pi_\theta(y_t | x, y_{<t}; \mathcal{R})}{\pi_{\text{old}}(y_t | x, y_{<t}; \mathcal{R})} \\
    &+ \mathbbm{1}\{t=|y|\} \cdot \max\{r\epsilon, r - \alpha c\}\nonumber
\end{align}

We use exact match for the final answer to compute the outcome-based reward value $r$, therefore $r \in \{0, 1\}$. Similar to the proposed approach for GRPO, the cost value $c$ can be computed in two different ways: a memory-bound cost that focuses on the output length (i.e., $c = |y|$) or a latency-bound cost that focuses on model's latency, for which we can reuse Equation~\eqref{eq:latency_cost} for computing $c$. The discounted reward value is bounded by zero (if the final answer is wrong) or $\epsilon$ (if the final answer is correct). This ensures that advantage function remains non-negative and regardless of the cost, there is at least a small positive reward (i.e., $\epsilon$) for a correct answer. We set $\epsilon$ to 0.2 in our experiments. The hyper-parameter $\alpha$ controls the impact of cost on the advantage function. As is common in PPO training \cite{PPO}, mean squared error (MSE) is used to optimize the model parameters for the value function $v_\phi$. Refer to \citet{ivison2024unpacking} for more details.

\section{Experiments}
\subsection{Data}
We run our experiments on seven diverse question answering datasets.  They range from general question answering datasets (usually single shot) to multi-hop question answering datasets that need reasoning and multiple rounds of retrieval. For general QA, we use Natural Questions, TriviaQA, and PopQA. For multi-hop QA, we use HotpotQA, 2wiki, Musique, and Bamboodgle. They are explained below. 

Natural Questions (NQ) \citep{kwiatkowski2019natural} is a large-scale, real-world benchmark for question answering that contains thousands of real user queries issued to the Google search engine. An annotator was given a question and a retrieved Wikipedia page and was asked to select a long answer (usually a paragraph) and a short answer (typically one or more entities) if they exist on the page. They were instructed to submit ``null'' if no suitable answer is found. The dataset is widely used to evaluate the effectiveness of extractive open-domain QA systems. 
NQ has ~307K training examples, ~7.8K dev, and ~7.8K test examples. Around 152,148 questions have a long answer and 110,724 of them have a short answer. Short answers can be sets of spans in the document (106,926), or yes or no (3,798). Long answers are HTML bounding boxes. 

TriviaQA \cite{joshi2017triviaqa} is a QA dataset that includes more than 650,000 question–answer–evidence triples. The question-answer pairs come  from 14 trivia and quiz-league websites. Each question-answer pairs come
with an average of six supporting evidence documents, with the corresponding evidence documents later gathered automatically from Wikipedia and the broader Web. Therefore, the supporting documents may not include all the information necessary to fully answer the questions. 
The information required to answer 40\% of questions in the dataset are scattered over multiple sentences. There are in total 95,956 question-answer pairs in the dataset. 
 
PopQA  \citep{mallen2022not} is a large-scale, entity-centric open-domain QA dataset that includes entities across a broad spectrum of popularity. To build PopQA, authors randomly selected knowledge triples representing 16 different types of relationships from Wikidata. Then they transformed these triples into natural language questions using manually crafted templates for each relationship type. Acceptable answers to each question are the set of entities such that (Subject, Relation, Entities) is present in the knowledge graph. The dataset contains 14k QA pairs with fine-grained Wikidata entity ID.

HotpotQA \citep{yang2018hotpotqa} is a large-scale, crowdsourced dataset designed to support multi-hop question answering over natural language text, specifically Wikipedia.  To construct the dataset, the authors utilized a Wikipedia hyperlink graph, focusing on links from the first paragraphs of articles. They identified ``bridge entities'' that connect different articles and serve as pivots for multi-hop reasoning. Additionally, HotpotQA introduces a novel class of comparison questions, where systems must compare two entities along shared attributes, such as determining who played for more teams or comparing numerical values like ages. These questions present new challenges that go beyond span extraction and require models to reason over multiple facts and perform basic arithmetic or semantic comparisons. The dataset also supports yes/no variants of comparison questions to further test reasoning depth. The dataset contains 112,779 example, 18,089 of which are single hop, and 56,814 of them are easy multi-hop questions that are used for training. The rest of the dataset contains hard multi-hop examples divided in dev and test sets. For experiments of this paper, we use the full-wiki part of dataset.

The 2WikiMultiHopQA dataset \citep{ho2020constructing}, provides comprehensive explanations by integrating both structured and unstructured data. To improve the explanation and evaluation of answers, they enrich each sample with additional information, including evidence that clearly and concisely supports the outputs. 
The dataset contains 12,576 training and 12,576 test instances.

To ensure that all questions require reasoning, Musique \citep{trivedi2022musique} proposed a bottom-up approach that carefully selects pairs of single-hop questions which are composable and interdependent; meaning that  one reasoning step crucially depends on the outcome of another. The test data has 1271 two-hop examples, 763 three-hop examples, 425 four-hop examples. 

Bamboogle \citep{press2022measuring} is a dataset of 125 manually crafted two-hop questions. It is designed to pose a high level of difficulty that a common internet search engine cannot answer its questions directly, even though both supporting evidence pieces are available on Wikipedia. The dataset is intended to evaluate how well a QA system can handle diverse compositional questions, though it offers less statistical robustness due to its size.

To ensure fair comparison and improve reproducibility, we use the pre-processing steps provided by the Search-R1 open-source repository.\footnote{\url{https://github.com/petergriffinjin/search-r1}} Therefore, all models and baselines share the same experimental setups and hyper-parameter setting strategies. It is important to note that following previous work \cite{jin2025searchr1}, we only use the training data from NQ and HotpotQA as representatives of general and multi-hop QA datasets. Therefore, the results presented on NQ and HotpotQA are considered as in-domain experiments and all other datasets are considered out-of-domain experiments.

\begin{table*}[t]
    \centering
        \caption{Main results. The best performance is set in bold. $^\dagger/^\star$ represents in-domain/out-domain datasets, meaning that only NQ and HotpotQA are used for training. Dynamic Search-R1 results are reported for the model with latency-bound cost penalization.}\label{tab:main}

    \footnotesize
    \resizebox{\textwidth}{!}{
    \begin{tabular}{lcccccccccc}
        \hline
        \multirow{2}{*}{\textbf{Model}} & \multicolumn{3}{c}{\textbf{General QA}} && \multicolumn{4}{c}{\textbf{Multi-Hop QA}} && \multirow{2}{*}{\textbf{Avg.}} \\
        \cline{2-4} \cline{6-9} 
         & \textbf{NQ$^\dagger$} & \textbf{TriviaQA$^\star$} & \textbf{PopQA$^\star$} && \textbf{HotpotQA$^\dagger$} & \textbf{2wiki$^\star$} & \textbf{Musique$^\star$} & \textbf{Bamboogle$^\star$} &&  \\
        \hline
        \multicolumn{8}{l}{\textbf{Qwen2.5-7B}} \\
        Direct Inference & 0.134 & 0.408 & 0.140 & & 0.183 & 0.250 & 0.031 & 0.120 && 0.181 \\
        CoT & 0.048 & 0.185 & 0.054 && 0.092 & 0.111 & 0.022 & 0.232 && 0.106 \\
        IRCoT & 0.224 & 0.478 & 0.301 && 0.133 & 0.149 & 0.072 & 0.224 && 0.239 \\
        Search-o1 & 0.151 & 0.443 & 0.131 && 0.187 & 0.176 & 0.058 & 0.296 && 0.206 \\
        RAG & 0.349 & 0.585 & 0.392 && 0.299 & 0.235 & 0.058 & 0.208 && 0.304 \\
        SFT & 0.318 & 0.354 & 0.121 && 0.217 & 0.259 & 0.066 & 0.112 && 0.207  \\
        R1 & 0.297 & 0.539 & 0.202 && 0.242 & 0.273 & 0.083 & 0.296 && 0.276  \\
        Search-R1 PPO & 0.480 & \textbf{0.638} & {0.457} && {0.433} & 0.382 & {0.196} & {0.432} && {0.431}  \\
        Search-R1 GRPO & 0.395 & 0.560 & 0.388 & & 0.326 & 0.297 & 0.125 & 0.360 & & 0.350 \\
        \hdashline
        Dynamic Search-R1 PPO & \textbf{0.493} & 0.627 & \textbf{0.470} && \textbf{0.468} & \textbf{0.386} & \textbf{0.212} & \textbf{0.435} && \textbf{0.442}\\
        Dynamic Search-R1 GRPO & 0.414 & 0.582 & 0.406 && 0.362 & 0.322 & 0.139 & 0.373 && 0.371 \\
        \hline
        \multicolumn{8}{l}{\textbf{Qwen2.5-3B}} \\
        Direct Inference & 0.106 & 0.288 & 0.108 && 0.149 & 0.244 & 0.020 & 0.024 && 0.134 \\
        CoT & 0.023 & 0.032 & 0.005 && 0.021 & 0.021 & 0.002 & 0.000 && 0.015 \\
        IRCoT & 0.111 & 0.312 & 0.200 && 0.164 & 0.171 & 0.067 & 0.240 && 0.181 \\
        Search-o1 & 0.238 & 0.472 & 0.262 && 0.221 & 0.218 & 0.054 & {0.320} && 0.255 \\
        RAG & 0.348 & 0.544 & 0.387 && 0.255 & 0.226 & 0.047 & 0.080 && 0.270  \\
        SFT & 0.249 & 0.292 & 0.104 && 0.186 & 0.248 & 0.044 & 0.112 && 0.176  \\
        R1 & 0.226 & 0.455 & 0.173 && 0.201 & 0.268 & 0.055 & 0.224 && 0.229  \\
        Search-R1 PPO & {0.406} & {0.587} & {0.435} && 0.284 & 0.273 & 0.049 & 0.088 & & 0.303  \\
        Search-R1 GRPO & 0.421 & 0.583 & 0.413 && 0.297 & 0.274 & 0.066 & 0.128 & & 0.312 \\
        \hdashline
        Dynamic Search-R1 PPO & 0.411 & \textbf{0.598} & \textbf{0.451} & & \textbf{0.317} & 0.292 & 0.085 & 0.116 && 0.324 \\
        Dynamic Search-R1 GRPO & \textbf{0.437} & \textbf{0.598} & 0.429 && 0.315 & \textbf{0.294} & \textbf{0.092} & \textbf{0.134} & & \textbf{0.328} \\
        \hline
    \end{tabular}
    }
\end{table*}

\begin{table*}[t]
    \centering
        \caption{Average response length (\# tokens) and latency (milliseconds) of models for each query on the Natural Questions datasets. Qwen-2.5-7b-base is used as the base LLM in all the experiments. PPO is used for all optimizations.}

    \begin{tabular}{llllllllllll}\hline
        & \multicolumn{3}{c}{\textbf{NQ}} && \multicolumn{3}{c}{\textbf{HotpotQA}} && \multicolumn{3}{c}{\textbf{Musique}} \\\cline{2-4} \cline{6-8} \cline{10-12}
        \textbf{Model} & \textbf{EM} & \textbf{\# tokens} & \textbf{latency} && \textbf{EM} & \textbf{\# tokens} & \textbf{latency} && \textbf{EM} & \textbf{\# tokens} & \textbf{latency} \\\hline
        Search-R1 & 0.480 & 1083 & 88.82 && 0.433 & 1273 &	96.29 && 0.196	& 1355 &	103.64\\
        Dynamic Search-R1 w/o cost penalty & 0.497 & 1387 & 102.98 && 0.469	& 1539	& 107.61 && 0.218 &	1592 & 115.36\\
        Dynamic Search-R1 - Memory-bound & 0.491 & 977 & 83.87 && 0.465 &	1064 &	91.10 && 0.215 &	1148	& 98.03\\
        Dynamic Search-R1 - Latency-bound & 0.493 & 991 & 70.43 && 0.468	& 1095 & 	82.86 && 0.212	& 1171 &	87.62\\
        \hline
    \end{tabular}
    \label{tab:latency}
\end{table*}

\subsection{Baselines} 
We use 9 baselines in our experiments. Most of them are retrieval-augmented methods and conduct reasoning. For a fair performance comparison, they all use the same base LLMs. The main and most competitive baseline for this work is search-R1 \cite{jin2025searchr1} with both PPO and GRPO training. The `Direct Inference'  baseline generates answers based on the LLM parametric knowledge with no reasoning and no access to a search engine. The Chain-of-Thought (CoT) \cite{wei2022chain} baseline conducts inference-time CoT reasoning, still with no access to a search engine.   IRCoT \cite{trivedi2022interleaving} is a retrieval-augmented approach for multi-step QA that proposes an interleaving solution, where the idea is to use retrieval to guide the chain-of-thought (CoT) and use CoT reasoning to guide the retrieval. Search-o1 \cite{searcho1} is a framework designed to enhance LLMs by incorporating an agent-based retrieval-augmented generation approach along with a specialized Reason-in-Documents module to refine retrieved content and eliminate noisy information. It embeds an agentic search process directly into the reasoning pipeline at inference with no training. 
RAG is the retrieval-augmentation approach proposed by \citet{lewis2020retrieval}.
SFT is a supervised fine-tuning baseline \cite{chung2022} that fine-tunes the LLM on various tasks paraphrased as instructions. R1 is a post-training approach based on the R1-Zero approach proposed by DeepSeek \cite{guo2025deepseek}. R1 also does not have access to search engine results. All models are trained on the same data. All retrieval-augmented models also use the same retrieval model, i.e., E5 \cite{wang2024textembeddingsweaklysupervisedcontrastive} which has shown strong out-of-distribution performance.

\begin{figure*}[t]
    \centering
    \begin{subfigure}[b]{0.49\textwidth}
    \centering
        \includegraphics[width=.7\textwidth]{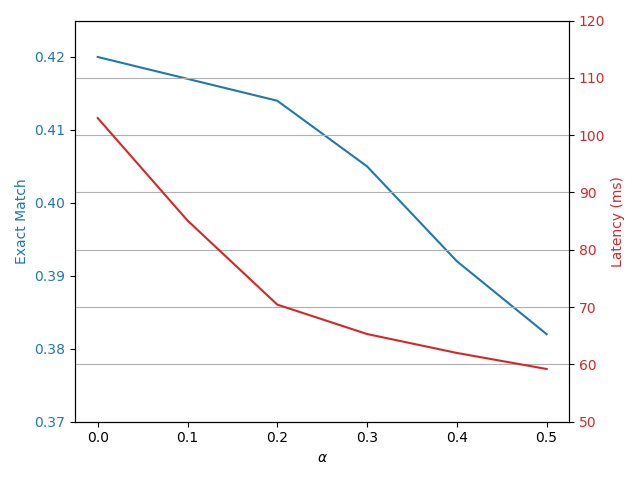}
        \caption{Latency-bound penalization}
        \label{fig:alpha1}
    \end{subfigure}
    \hfill
    \begin{subfigure}[b]{0.49\textwidth}
    \centering
        \includegraphics[width=.7\textwidth]{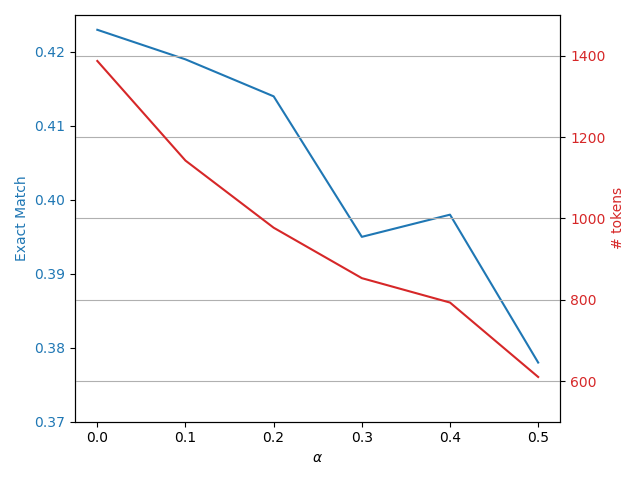}
        \caption{Memory-bound penalization}
        \label{fig:alpha2}
    \end{subfigure}
    
    \caption{The impact of parameter $\alpha$ on Dynamic Search-R1 GRPO with latency-bound cost penalization. See the plots in color.}
    \label{fig:alpha}
\end{figure*}

\subsection{Results and Discussion}

\paragraph{Comparison with the Baselines.} Table~\ref{tab:main} contains results obtained by our cost-aware (latency-bound) optimization approach based on Search-R1 \cite{jin2025searchr1} as well as the baselines. To have a fair comparison, we made sure that all models use the same base language model. Following \citet{jin2025searchr1}, we experiment with two model sizes: Qwen 2.5 \cite{qwen2025qwen25technicalreport} with 7 billion (top) and 3 billion parameters (bottom). The results are in terms of exact match and the last column reports the macro average performance across all seven datasets. As expected, the results obtained by models with the 7B LLM are substantially higher than the ones obtained by the 3B LLM. It is important to highlight that these improvements are more significant for the multi-hop QA datasets that are more challenging and require reasoning. The results also suggest that there is no winner among PPO and GRPO for Dynamic Search-R1; for the 7B LLM, our PPO-based model performs better than the one optimized by GRPO, while GRPO shows a small advantage when the LLM has 3 billion parameters. That said, the average GRPO improvement on 3B is relatively marginal, while PPO improvement on the 7B LLM is substantial. This is also inline with findings in Search-R1 \cite{jin2025searchr1}. We believe that it takes a model with relatively high capacity to benefit from a trained value function as is done in PPO. Dynamic Search-R1 outperforms all the baselines (including Search-R1) on all datasets, except for TriviaQA for the 7 billion base LLM. The main reason for observed improvements over Search-R1 is the ability of Dynamic Search-R1 in asking for more information from the retrieval results. Another reason may be due to the cost-aware objective, which discourages the model from repeating the same queries in case of retrieval failure (which happens quite frequently in Search-R1) and changing the query. As mentioned earlier, Dynamic Search-R1 results on this dataset are obtained based on our cost-aware optimization with latency-bound assumption, which is even more efficient than Search-R1 (as we will see in the next experiment).

\begin{table*}[h]
    \centering
    \caption{Two case studies demonstrating latency-bound Dynamic Search-R1 model with a 7B Qwen 2.5 base LLM. Both of these examples are also studied in the Search-R1 paper \cite{jin2025searchr1}.}

    \resizebox{\textwidth}{!}{
    \begin{tabular}{p{\textwidth}}
        \hline
        \textbf{Question}: What type of profession does Chris Jericho and Gary Barlow have in common? \\
        \hline
        \textbf{Ground-truth Answer}: musician \\
        \hline
        \think{I need to find the profession that Chris Jericho and Gary Barlow have in common.} \\
\search{Chris Jericho and Gary Barlow common profession} \\
\info{Doc 2(Title: "Chris Jericho") ""Do You Wanna Start a War"" (2014) and ""Judas"" (2017) are comprised entirely or predominantly of original compositions. He has also appeared on numerous TV shows over the years, having competed in the 2011 season of ""Dancing With the Stars"", lasting until the sixth week. Prior to that, Jericho hosted the ABC game show ""Downfall"" and the Revolver Golden Gods Awards. Christopher Keith Irvine was born on November 9, 1970, in Manhasset, New York, but grew up in Winnipeg. His father, ice hockey player Ted Irvine, had been playing for the New York Rangers at the time of his birth. ...
} \\

\think{Chris Jericho is a professional wrestler, musician, author, and actor. Now I need to find if Gary Barlow has any of these professions in common.} \\
\search{Gary Barlow profession} \\
\info{Doc 1(Title: "Gary Barlow") hits, twelve number one singles and seven number one albums with Take That. Barlow has received six Ivor Novello Awards from the British Academy of Songwriters, Composers and Authors, including the award for Outstanding Services to British Music. He has sold over 50 million records worldwide. Barlow was born on 20 January 1971 in Frodsham, Cheshire, the second son of Colin (died 15 October 2009) and Marjorie Barlow. In Barlow's autobiography, he relates that his love of music began at an early age, and calls himself ""one of those kids that's forever dancing in front of the TV looking at ...
} \\
\think{Gary Barlow is a singer, songwriter, musician, and record producer. The common profession between Chris Jericho and Gary Barlow is musician.} \\
\answer{musician}\\
        \hline
        \hline
        \textbf{Question}: Where are Teide National Park and Garajonay National Park located? \\
        \hline
        \textbf{Ground-truth Answer}: Canary Islands, Spain \\
        \hline
        \think{I need to find out where Teide National Park and Garajonay National Park are located.} \\
\search{Teide National Park and Garajonay National Park location} \\
\info{Doc 1(Title: "Garajonay National Park") decided to take their own lives. Garajonay National Park Garajonay National Park (, ) is located in the center and north of the island of La Gomera, one of the Canary Islands (Spain). It was declared a national park in 1981 and a World Heritage Site by UNESCO in 1986. It occupies 40 km (15 sq mi) and it extends into each of the six municipalities on the island. Nearly 450,000 tourists visit the park every year. There is accommodation for tourists just outside the park's boundary. The park is named after the rock formation of Garajonay, the highest point ...
} \\
\think{Garajonay National Park is located in the center and north of the island of La Gomera in Canary Islands, Spain. I need more information to learn about Teide National Park.} \\
\more{1} \\
\info{Doc 1(Title: "Teide National Park") Teide National Park Teide National Park (, ) is a national park located in Tenerife (Canary Islands, Spain). The national park is centered on Mount Teide, the highest mountain of Spain (3,718 meters high) in addition, the highest volcano in the Atlantic Ocean islands and the third highest volcano in the world from its base in the ocean floor (7.500 meters high). Its national park status was passed on 22 January 1954, making it the third oldest national park in Spain. Pico Viejo, also included within the national park limits, is the second highest volcano in the Canary Islands with ...
} \\
\think{I found out that Teide National Park is located in Tenerife in Canary Islands, Spain. They are both located in Canary Islands.} \\
\answer{Canary Islands, Spain}\\
\hline
\end{tabular}
}
\label{tab:case_study}
\end{table*}

\paragraph{The Impact of Cost Penalization.} In the next set of experiments, we evaluate the impact of cost-awareness methods on Dynamic Search-R1 performance. For a fair comparison, we use one A100 GPU with 80 GB memory for this experiment and all models are based on the Qwen 2.5 model with 7 billion parameters. For the sake of space, we conduct this ablation study on the PPO algorithm and for only three of our datasets; we select NQ, HotpotQA, and Musique to cover diverse datasets. That said, we made similar observations on all datasets.  According to the results presented in Table~\ref{tab:latency}, Dynamic Search-R1 without any efficiency-aware penalization produces more tokens and is slower than Search-R1. This happens because Dynamic Search-R1 keeps asking for more information (i.e., more retrieved documents) when it cannot find the information it needs in the retrieval results. We observe that the latency gap is much larger for the queries for which the model cannot generate the correct answer, which indicates the model's trial for getting relevant information as much as the budget $B$ (see Algorithm~\ref{alg:llm_search}) allows. Both memory- and latency-bound approaches significantly improve efficiency metrics. The memory-bound method leads to the minimum number of tokens. The latency-bound method performs faster, even though it produces slightly more tokens compared to the memory-bound alternative. The reason for this observation is that the latency-bound model generates fewer tokens and there are more retrieved tokens in its reasoning chain. Therefore, since token generation is more expensive and time-consuming than token encoding, the latency-bound approach leads to lower end-to-end latency. Note that all Dynamic Search-R1 methods perform comparably on the Natural Questions dataset, demonstrating that cost-aware penalization improves efficiency without compromising effectiveness.

To better understand the impact of the parameter $\alpha$ on the efficiency and effectiveness of Dynamic Search-R1, we compute average exact match and latency for the Natural Questions dataset. The curves are depicted in Figure~\ref{fig:alpha1}. Without loss of generality, this experiment only focuses on GRPO optimization and with Qwen 2.5 7B as the base LLM. According to the figure, increasing $\alpha$ would lead to a decrease in Exact Match and Latency, meaning that higher $\alpha$ values hurt effectiveness but improves efficiency, that said the performance drop for large enough $\alpha$ would be significant which may not be desired. System designers can choose the right value of $\alpha$ based on their applications. In the main results presented in this paper, we choose the highest value of alpha that does not lead to statistically significant ($p\_value<0.01$ with McNemar test) performance drop compared to $\alpha=0$ on the validation set. That resulted in the empirical value of $\alpha=0.2$. 
We also plot the impact of $\alpha$ values on the GRPO model with memory-bound penalization in Figure~\ref{fig:alpha2}. The observation is relatively similar to that of the latency-bound model.

\section{Case Studies}
\label{appendix:case_studies}
We intentionally choose two questions presented in the Search-R1 paper \cite{jin2025searchr1} that would highlight how Cost-aware Dynamic Search-R1 performs. Two case studies for the latency-bound 7B model trained with PPO is presented in Table~\ref{tab:case_study}. Both of them are selected from the HotPotQA dataset. For the first example in the table, our method produces less than half the number of tokens Search-R1 produces (compare with Table 10 in \cite{jin2025searchr1}). In the second example, it uses the same number of retrieved documents as Search-R1 (see Table 13 in \cite{jin2025searchr1}), but generates fewer tokens by shortening the reasoning chain and asking for more information instead of submitting a second search query.

\section{Conclusions and Future Work}
This work introduced Dynamic Search-R1--a retrieval-augmented reasoning model that enables rich interactions between reasoning-powered LLMs and search engines, such as query generation and asking for more documents deeper in the ranked list. It also explored cost-aware solutions for training Dynamic Search-R1 models that are efficient by introducing both memory- and latency-bound penalization functions. Extensive experiments on seven public datasets demonstrate the effectiveness and efficiency of the proposed models in various settings, highlighting significant improvements over competitive baselines while performing faster with fewer reasoning tokens. 

In the future, we plan to apply this work to the broader agentic AI models, where a cost is associated with each agent and the optimization focuses on solving complex tasks with the lowest cost possible. We also plan to extend our cost-aware approach to multi-modal tasks and further go beyond memory- and latency-bound cost penalization, for example, by considering energy consumption and CO2 emission.

\bibliographystyle{ACM-Reference-Format}
\bibliography{sample-base}











\end{document}